\documentclass[11pt]{article}

\usepackage{lineno,hyperref}
\modulolinenumbers[5]

\usepackage{amsfonts}
\usepackage{amsmath}
\usepackage{listings}
\usepackage{algorithm}
\usepackage{algorithmic}
\usepackage{booktabs}
\usepackage{gensymb}
\usepackage{subcaption}
\usepackage[toc,page]{appendix}
\usepackage{xcolor}
\usepackage{graphicx}
\usepackage{arxiv}
\usepackage{setspace}

\title{Gym-ANM: Open-source software to leverage reinforcement learning for power system management in research and education}

\date{}
\author{
  Robin~Henry \\
  The University of Edinburgh\\
  \texttt{robin@robinxhenry.com} \\
   \And
 Damien~Ernst \\
  The University of Liège \\
  \texttt{dernst@uliege.be} \\
}

\begin{document}
\maketitle

\begin{abstract}
Gym-ANM is a Python package that facilitates the design of reinforcement learning (RL) environments that model active network management (ANM) tasks in electricity networks. Here, we describe how to implement new environments and how to write code to interact with pre-existing ones. We also provide an overview of ANM6-Easy, an environment designed to highlight common ANM challenges. Finally, we discuss the potential impact of Gym-ANM on the scientific community, both in terms of research and education. We hope this package will facilitate collaboration between the power system and RL communities in the search for algorithms to control future energy systems.
\end{abstract}

\keywords{Gym-ANM \and reinforcement learning \and active network management \and distribution networks \and renewable energy}


\section{Introduction}
Active network management (ANM) of electricity distribution networks is the process of controlling generators, loads, and storage devices for specific purposes (e.g., minimizing operating costs, keeping voltages and currents within operating limits) \cite{gill2013dynamic}. The modernization of distribution networks is taking place with the addition of distributed renewable energy resources and storage devices. This attempt to transition towards sustainable energy systems leaves distribution network operators (DNO) facing many new complex ANM problems (overvoltages, transmission line congestion, voltage coordination, investment issues, etc.) \cite{mcdonald2008adaptive}. 

There is a growing belief that reinforcement learning (RL) algorithms have the potential to tackle these complex ANM challenges more efficiently than traditional optimization methods. This optimism results from the fact that RL approaches have been successfully and extensively applied to a wide range of fields with similarly difficult decision-making problems, including games \cite{mnih2013playing, mnih2015human, silver2016mastering, vinyals2019grandmaster}, robotics \cite{deisenroth2013survey, kormushev2013reinforcement, kober2013reinforcement, gu2017deep}, and autonomous driving \cite{sallab2017deep, o2018scalable, li2019reinforcement}.

What games, robotics, and autonomous driving all have in common is that the environment in which the decisions have to be taken can be efficiently replicated using open-source software simulators. In addition, these software libraries usually provide interfaces tailored for writing code for RL research. Hence, the availability of such packages makes it easier for RL researchers to apply their algorithms to decision-making problems in these fields, without needing to first develop a deep understanding of the underlying dynamics of the environments with which their agents interact. 

Put simply, we believe that ANM-related problems would benefit from a similar amount of attention from the RL community if open-source software simulators were available to model them and provide a simple interface for writing RL research code. With that in mind, we designed Gym-ANM, an open-source Python package that facilitates the design and the implementation of RL environments that model ANM tasks \cite{henry2021gym}. Its key features, which differentiate it from traditional power system modeling software (e.g., MATPOWER \cite{zimmerman2010matpower}, pandapower \cite{pandapower}), are:
\begin{itemize}
    \item Very little background in power system modeling is required, since most of the complex dynamics are abstracted away from the user.
    \item The environments (tasks) built using Gym-ANM follow the OpenAI Gym interface \cite{brockman2016openai}, with which a large part of the RL community is already familiar.
    \item The flexibility of Gym-ANM, with its different customizable components, makes it a suitable framework to model a wide range of ANM tasks, from simple ones that can be used for educational purposes, to complex ones designed to conduct advanced research. 
\end{itemize}

Finally, as an example of the type of environment that can be built using Gym-ANM, we also released ANM6-Easy, an environment that highlights common ANM challenges in a 6-bus distribution network. 

Both the Gym-ANM framework and the ANM6-Easy environment, including detailed mathematical formulations, were previously introduced in \cite{henry2021gym}. Here, our goal is to provide a short practical guide to the use of the package and discuss the impact that it may have on the research community.

\section{The Gym-ANM package}

The Gym-ANM package was designed to be used for two particular use cases. The first is the design of novel environments (ANM tasks), which requires writing code that simulates generation and demand curves for each device connected to the power grid (Section \ref{sec:design-env}). The second use case is the training of RL algorithms on an existing environment (Section \ref{sec:use-env}).

\subsection{Design a Gym-ANM environment}
\label{sec:design-env}

\begin{figure}[h!]
    \centering
    \includegraphics[width=\textwidth]{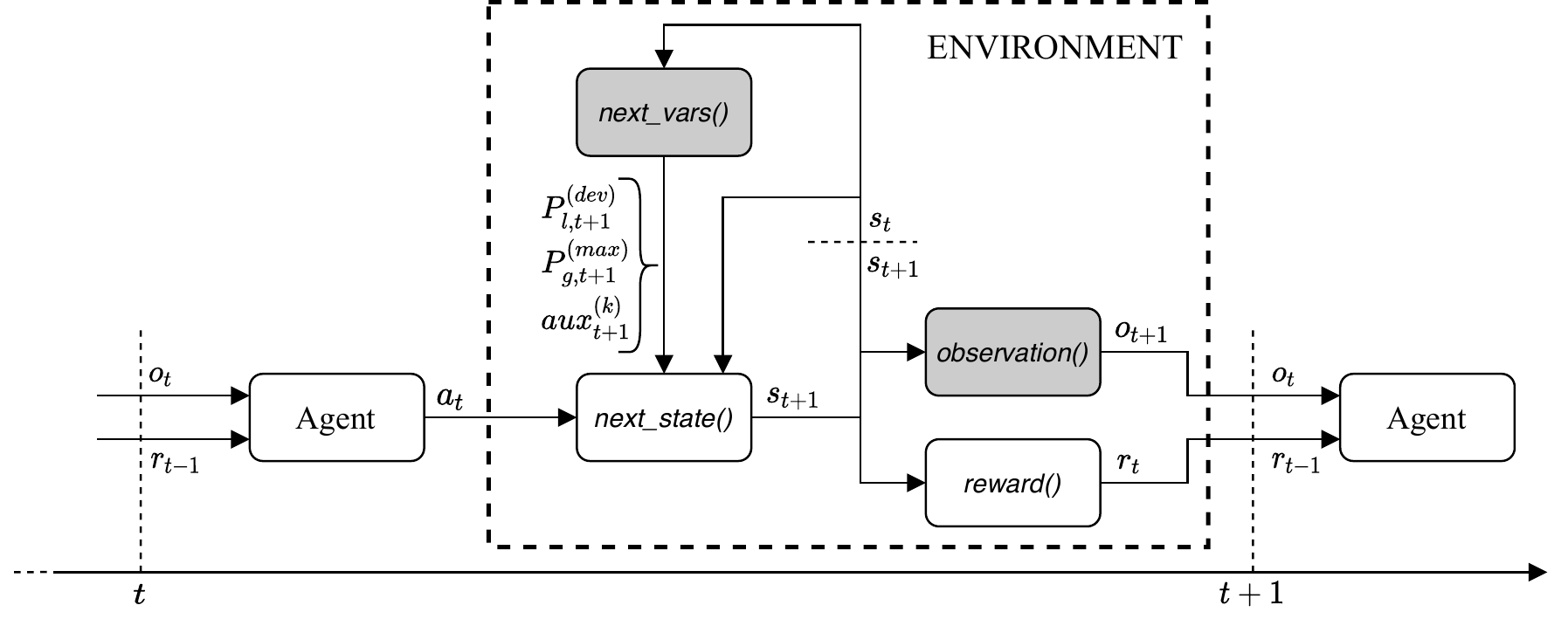}
    \caption{Internal structure of a Gym-ANM environment, taken from \cite{henry2021gym}.}
    \label{fig:environment-structure}
\end{figure}

The internal structure of a Gym-ANM environment is shown in Figure \ref{fig:environment-structure}. At each timestep, the agent passes an action $a_t$ to the environment. The latter generates a set of stochastic variables by calling the \texttt{next\_vars()} function, which are then used along with $a_t$ to simulate the distribution network and transition to a new state $s_{t+1}$. Finally, the environment outputs an observation vector $o_{t+1}$ and a reward $r_t$ through the \texttt{observation()} and \texttt{reward()} functions. 

The core of the power system modeling is abstracted from the user in the \texttt{next\_state()} call. The grey blocks, \texttt{next\_vars()} and \texttt{observation()}, are the only components that are fully customizable when designing new Gym-ANM environments.

In practice, new environments are created by implementing a sub-class of \texttt{ANMEnv}. The general template to follow is shown in Listing \ref{lst:new_env}. A more detailed description, along with examples, can be found in the online documentation\footnote{\url{https://gym-anm.readthedocs.io/en/latest/topics/design_new_env.html}}. 

\clearpage
{\footnotesize
\begin{lstlisting}[language=Python, caption={Implementation template for new Gym-ANM environments.}, label=lst:new_env, captionpos=b, frame=single, escapeinside={(*}{*)}, xleftmargin=3.4pt, xrightmargin=3.4pt]
from gym_anm import ANMEnv

class CustomEnvironment(ANMEnv):
    def __init__(self):
        network = {'baseMVA': ..., 'bus': ...,
                   'device': ..., 'branch': ...}  
                             # power grid specs
        observation = ...    # observation space
        K = ...              # number of auxiliary variables
        delta_t = ...        # timestep intervals
        gamma = ...          # discount factor
        lamb = ...           # penalty hyperparameter
        aux_bounds = ...     # bounds on auxiliary variable
        costs_clipping = ... # reward clipping parameters
        seed = ...           # random seed

        super().__init__(network, observation, K, delta_t, 
                         gamma, lamb, aux_bounds, 
                         costs_clipping, seed)
    
    # Return an initial state vector (*$s_0 \sim p_0(\cdot)$*).
    def init_state(self):
        ...
    # Return the next stochastic variables. 
    def next_vars(self, s_t):
        ...
    # Return the bounds of the observation vector space. 
    def observation_bounds(self):  # optional
        ...
\end{lstlisting}}

\subsection{Use a Gym-ANM environment}
\label{sec:use-env}
A code snippet illustrating how a custom Gym-ANM environment can be used alongside an RL agent implementation is shown in Listing \ref{lst:code_snippet}. Note that for clarity, this example omits the agent-learning procedure. Because Gym-ANM is built on top of the Gym toolkit \cite{brockman2016openai}, all Gym-ANM environments provide the same interface as traditional Gym environments, as described in their online documentation\footnote{\url{https://gym.openai.com/docs/}}.

{\footnotesize
\begin{lstlisting}[language=Python, caption={A Python code snippet illustrating environment-agent interactions \cite{henry2021gym}.}, label=lst:code_snippet, captionpos=b, escapeinside={(*}{*)}]
env = gym.make('MyANMEnv')   # Initialize the environment.
obs = env.reset()            # Reset the env. and collect (*$o_0$*).

for t in range(1, T):         
  env.render()        # Update the rendering.
  a = agent.act(obs)  # Agent takes (*$o_t$*) as input and chooses (*$a_t$*).
  obs, r, done, info = env.step(a) 
            # The action (*$a_t$*) is applied, and are outputted:
            # - obs: the new observation (*$o_{t+1}$*),
            # - r: the reward (*$r(s_t, a_t, s_{t+1})$*),
            # - done: True if (*$s_{t+1} \in \mathcal S^{terminal}$*),
            # - info: extra info about the transition.

env.close() # Close the environment and stop rendering.
\end{lstlisting}} 

\section{Example: the ANM6-Easy environment}
\begin{figure}[h!]
    \centering
    \includegraphics[width=\textwidth]{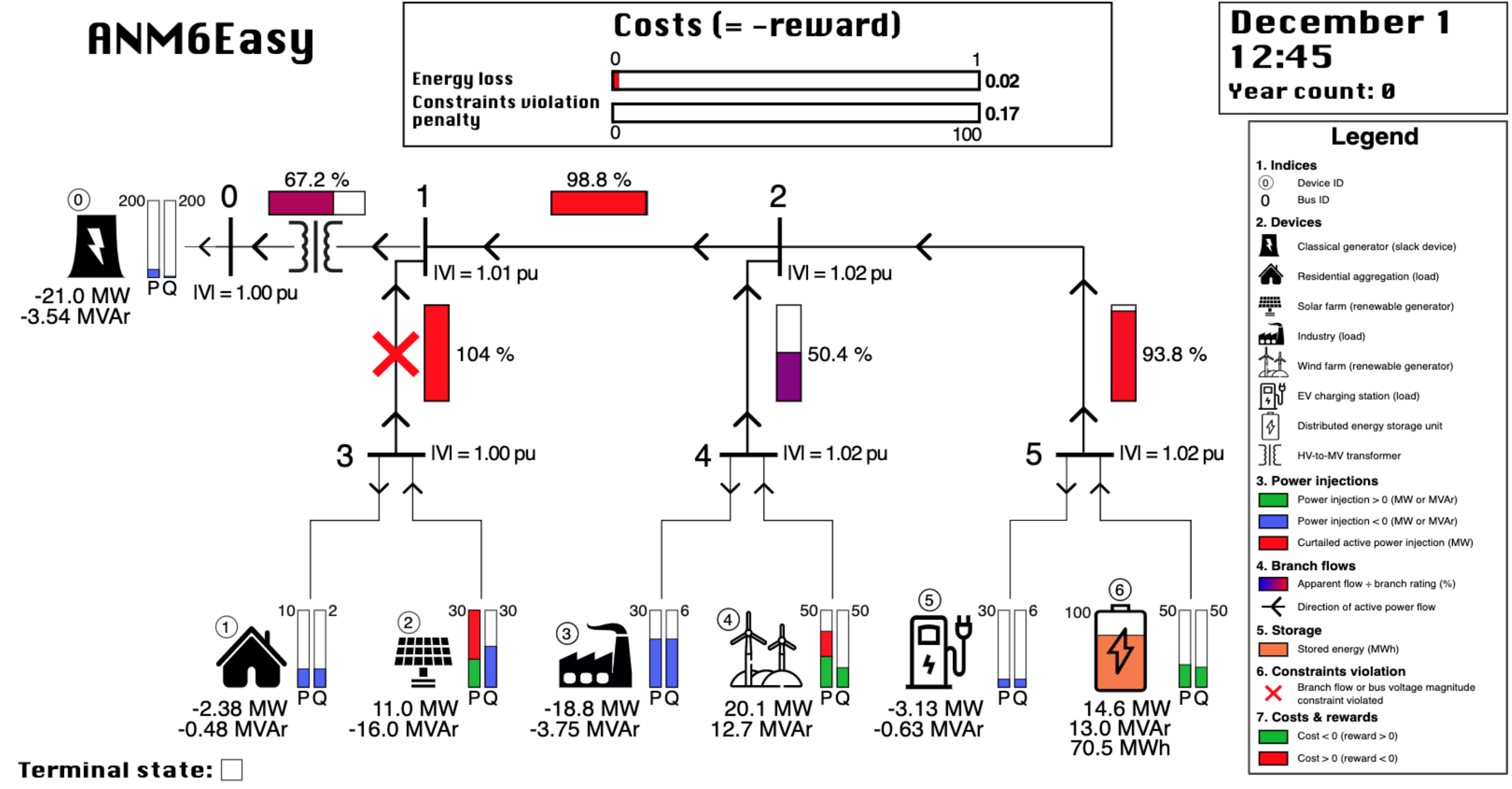}
    \caption{The ANM6-Easy Gym-ANM environment, taken from \cite{henry2021gym}.}
    \label{fig:ANM6-Easy}
\end{figure}

ANM6-Easy is the first Gym-ANM environment that we have released \cite{henry2021gym}. It models a 6-bus network and was engineered so as to highlight some of the most common ANM challenges faced by network operators. A screenshot of the rendering of the environment is shown in Figure \ref{fig:ANM6-Easy}. 

In order to limit the complexity of the task, the environment was designed to be fully deterministic: both the demand from loads (1: residential area, 3: industrial complex, 5: EV charging garage) and the maximum generation (before curtailment) profiles from the renewable energies (2: solar farm, 4: wind farm) are modelled as fixed 24-hour time series that repeat every day, indefinitely. 

More information about the ANM6-Easy environment can be found in the online documentation\footnote{\url{https://gym-anm.readthedocs.io/en/latest/topics/anm6_easy.html}}.

\section{Research and educational impact}

Many software applications exist for modeling steady-state power systems in industrial settings, such as PowerFactory \cite{gonzalez2014powerfactory}, ERACS \cite{langley1992eracs}, ETAP \cite{brown1990interactive}, IPSA \cite{AboutIps54}, and PowerWorld \cite{PowerWor64}, all of which require a paid license. In addition, these programs are not well suited to conduct RL research since they do not integrate well with the two programming languages mostly used by the RL community: MATLAB and Python. Among the power system software packages that do not require an additional license and that are compatible with these programming languages, the commonly used in power system management research are MATPOWER (MATLAB) \cite{zimmerman2010matpower}, PSAT (MATLAB) \cite{milano2005open}, PYPOWER (Python interface for MATPOWER) \cite{rwlPYPOW4}, and pandapower (Python) \cite{pandapower}. 

Nevertheless, using the aforementioned software libraries to design RL environments that model ANM tasks is not ideal. First, the user needs to become familiar with the modeling language of the library, which already requires a good understanding of the inner workings of the various components making up power systems and of their interactions. Second, these packages often include a large number of advanced features, which is likely to overwhelm the inexperienced user and get in the way of designing even simple ANM scenarios. Third, because these libraries were designed to facilitate a wide range of simulations and analyses, they often do so at the cost of solving simpler problems more slowly (e.g., simple AC load flows). Fourth, in the absence of a programming framework agreed upon by the RL research community interested in tackling energy system management problems, various research teams are likely to spend time and resources implementing the same underlying dynamics common to all such problems. 

By releasing Gym-ANM, we hope to address all the shortcomings of traditional modeling packages described in the previous paragraph. Specifically:
\begin{itemize}
    
    \item The dissociation between the design of the environment (Section \ref{sec:design-env}) and the training of RL agents on it (Section \ref{sec:use-env}) encourages collaboration between researchers experienced in power system modeling and in RL algorithms. Thanks to the general framework provided by Gym-ANM, each researcher may focus on their particular area of expertise (designing or solving the environment), without having to worry about coordinating their implementations. 
    
    \item This dissociation also means that RL researchers are able to tackle the ANM tasks modelled by Gym-ANM environments without having to first understand the complex dynamics of the system. As a result, existing Gym-ANM environments can be explored by many in the RL community, from novices to experienced researchers. This is further facilitated by the fact that all Gym-ANM environments implement the Gym interface, which allows RL users to apply their own algorithms to any Gym-ANM task with little code modification (assuming they have used Gym in the past). 
    
    \item Gym-ANM focuses on a particular subset of ANM problems. This specificity has two advantages. The first is that it simplifies the process of designing new environments, since only a few components need to be implemented by the user. The second is that, during the implementation of the package, it allowed us to focus on simplicity and speed. That is, rather than providing a large range of modeling features like most of the other packages, we focused on optimizing the computational steps behind the \texttt{next\_state()} block of Figure \ref{fig:environment-structure} (i.e., solving AC load flows). This effectively reduces the computational time required to train RL agents on environments built with Gym-ANM. 
    
\end{itemize}

The simplicity with which Gym-ANM can be used by both the power system modeling and the RL communities has an additional advantage: it makes it a great teaching tool. This is particularly true for individuals interested in working at the intersection of power system management and RL research. One of the authors, Damien Ernst, has recently started incorporating the ANM6-Easy task in his RL course, \textit{Optimal decision making for complex systems}, at the University of Liège \cite{INFO800395}. 

Finally, we also compared the performance of the soft actor-critic (SAC) and proximal policy optimization (PPO) RL algorithms against that of an optimal model predictive control (MPC) policy on the ANM6-Easy task in \cite{henry2021gym}. We showed that, with almost no hyperparameter tuning, the RL policies were already able to reach near-optimal performance. These results suggest that state-of-the-art RL methods have the potential to compete with, or even outperform, traditional optimization approaches in the management of electricity distribution networks. Of course, ANM6-Easy is only a toy example, and confirming this hypothesis will require the design of more complex and advanced Gym-ANM environments.

\section{Conclusions and future works}

In this paper, we discussed the usage of the Gym-ANM software package first introduced in \cite{henry2021gym}, as well as its potential impact on the research community. We created Gym-ANM as a framework for the RL and energy system management communities to collaborate on tackling ANM problems in electricity distribution networks. As such, we hope to contribute to the gathering of momentum around the applications of RL techniques to challenges slowing down the transition towards more sustainable energy systems. 

In the future, we plan to design and release Gym-ANM environments that more accurately model real-world distribution networks as opposed to that modeled by ANM6-Easy. However, we also highly encourage other teams to design and release their own Gym-ANM tasks and/or to attempt to solve existing ones. 

\section*{Declaration of competing interests}
The authors declare that they have no known competing financial interests or personal relationships that could have appeared to influence the work reported in this paper.

\section*{Acknowledgements}
We would like to thank Raphael Fonteneau, Quentin Gemine, and Sébastien Mathieu at the University of Liège for their valuable early feedback and advice, as well as Gaspard Lambrechts and Bardhyl Miftari for the feedback they provided as the first users of Gym-ANM.

\bibliography{references}

\begin{thebibliography}{10}

\bibitem{gill2013dynamic}
Simon Gill, Ivana Kockar, and Graham~W Ault.
\newblock Dynamic optimal power flow for active distribution networks.
\newblock {\em IEEE Transactions on Power Systems}, 29(1):121--131, 2013.

\bibitem{mcdonald2008adaptive}
Jim McDonald.
\newblock Adaptive intelligent power systems: Active distribution networks.
\newblock {\em Energy Policy}, 36(12):4346--4351, 2008.

\bibitem{mnih2013playing}
Volodymyr Mnih, Koray Kavukcuoglu, David Silver, Alex Graves, Ioannis
  Antonoglou, Daan Wierstra, and Martin Riedmiller.
\newblock Playing {A}tari with deep reinforcement learning.
\newblock {\em arXiv preprint arXiv:1312.5602}, 2013.

\bibitem{mnih2015human}
Volodymyr Mnih, Koray Kavukcuoglu, David Silver, Andrei~A Rusu, Joel Veness,
  Marc~G Bellemare, Alex Graves, Martin Riedmiller, Andreas~K Fidjeland, Georg
  Ostrovski, et~al.
\newblock Human-level control through deep reinforcement learning.
\newblock {\em Nature}, 518(7540):529--533, 2015.

\bibitem{silver2016mastering}
David Silver, Aja Huang, Chris~J Maddison, Arthur Guez, Laurent Sifre, George
  Van Den~Driessche, Julian Schrittwieser, Ioannis Antonoglou, Veda
  Panneershelvam, Marc Lanctot, et~al.
\newblock Mastering the game of {G}o with deep neural networks and tree search.
\newblock {\em Nature}, 529(7587):484, 2016.

\bibitem{vinyals2019grandmaster}
Oriol Vinyals, Igor Babuschkin, Wojciech~M Czarnecki, Micha{\"e}l Mathieu,
  Andrew Dudzik, Junyoung Chung, David~H Choi, Richard Powell, Timo Ewalds,
  Petko Georgiev, et~al.
\newblock Grandmaster level in starcraft ii using multi-agent reinforcement
  learning.
\newblock {\em Nature}, 575(7782):350--354, 2019.

\bibitem{deisenroth2013survey}
Marc~Peter Deisenroth, Gerhard Neumann, Jan Peters, et~al.
\newblock A survey on policy search for robotics.
\newblock {\em Foundations and Trends{\textregistered} in Robotics},
  2(1--2):1--142, 2013.

\bibitem{kormushev2013reinforcement}
Petar Kormushev, Sylvain Calinon, and Darwin~G Caldwell.
\newblock Reinforcement learning in robotics: Applications and real-world
  challenges.
\newblock {\em Robotics}, 2(3):122--148, 2013.

\bibitem{kober2013reinforcement}
Jens Kober, J~Andrew Bagnell, and Jan Peters.
\newblock Reinforcement learning in robotics: A survey.
\newblock {\em The International Journal of Robotics Research},
  32(11):1238--1274, 2013.

\bibitem{gu2017deep}
Shixiang Gu, Ethan Holly, Timothy Lillicrap, and Sergey Levine.
\newblock Deep reinforcement learning for robotic manipulation with
  asynchronous off-policy updates.
\newblock In {\em 2017 IEEE international conference on robotics and automation
  (ICRA)}, pages 3389--3396. IEEE, 2017.

\bibitem{sallab2017deep}
Ahmad~EL Sallab, Mohammed Abdou, Etienne Perot, and Senthil Yogamani.
\newblock Deep reinforcement learning framework for autonomous driving.
\newblock {\em Electronic Imaging}, 2017(19):70--76, 2017.

\bibitem{o2018scalable}
Matthew O'Kelly, Aman Sinha, Hongseok Namkoong, Russ Tedrake, and John~C Duchi.
\newblock Scalable end-to-end autonomous vehicle testing via rare-event
  simulation.
\newblock In {\em Advances in Neural Information Processing Systems}, pages
  9827--9838, 2018.

\bibitem{li2019reinforcement}
Dong Li, Dongbin Zhao, Qichao Zhang, and Yaran Chen.
\newblock Reinforcement learning and deep learning based lateral control for
  autonomous driving [application notes].
\newblock {\em IEEE Computational Intelligence Magazine}, 14(2):83--98, 2019.

\bibitem{henry2021gym}
Robin Henry and Damien Ernst.
\newblock Gym-anm: Reinforcement learning environments for active network
  management tasks in electricity distribution systems.
\newblock {\em arXiv preprint arXiv:2103.07932}, 2021.

\bibitem{zimmerman2010matpower}
Ray~Daniel Zimmerman, Carlos~Edmundo Murillo-S{\'a}nchez, and Robert~John
  Thomas.
\newblock Matpower: Steady-state operations, planning, and analysis tools for
  power systems research and education.
\newblock {\em IEEE Transactions on power systems}, 26(1):12--19, 2010.

\bibitem{pandapower}
L.~Thurner, A.~Scheidler, F.~Sch{\"a}fer, J.~Menke, J.~Dollichon, F.~Meier,
  S.~Meinecke, and M.~Braun.
\newblock pandapower — an open-source python tool for convenient modeling,
  analysis, and optimization of electric power systems.
\newblock {\em IEEE Transactions on Power Systems}, 33(6):6510--6521, Nov 2018.

\bibitem{brockman2016openai}
Greg Brockman, Vicki Cheung, Ludwig Pettersson, Jonas Schneider, John Schulman,
  Jie Tang, and Wojciech Zaremba.
\newblock Openai {G}ym.
\newblock {\em arXiv preprint arXiv:1606.01540}, 2016.

\bibitem{gonzalez2014powerfactory}
Francisco~M Gonzalez-Longatt and Jos{\'e}~Luis Rueda.
\newblock {\em PowerFactory applications for power system analysis}.
\newblock Springer, 2014.

\bibitem{langley1992eracs}
HJ~Langley and KD~Wright.
\newblock Eracs-a comprehensive package for pcs.
\newblock In {\em IEE Colloquium on Interactive Graphic Power System Analysis
  Programs}, pages 3--1. IET, 1992.

\bibitem{brown1990interactive}
Keith Brown, Farrokh Shokooh, Herminio Abcede, and Gary Donner.
\newblock Interactive simulation of power systems: Etap applications and
  techniques.
\newblock In {\em Conference Record of the 1990 IEEE Industry Applications
  Society Annual Meeting}, pages 1930--1941. IEEE, 1990.

\bibitem{AboutIps54}
{TNEI}.
\newblock {Interactive Power System Analysis} ({IPSA}) software.
\newblock \url{https://www.ipsa-power.com}.
\newblock (Accessed on 05/12/2021).

\bibitem{PowerWor64}
{PowerWorld Corporation}.
\newblock {PowerWorld} software.
\newblock \url{https://www.powerworld.com}.
\newblock (Accessed on 05/12/2021).

\bibitem{milano2005open}
Federico Milano.
\newblock An open source power system analysis toolbox.
\newblock {\em IEEE Transactions on Power systems}, 20(3):1199--1206, 2005.

\bibitem{rwlPYPOW4}
Richard Lincoln.
\newblock {PYPOWER} library.
\newblock \url{https://github.com/rwl/PYPOWER}.
\newblock (Accessed on 05/12/2021).

\bibitem{INFO800395}
Damien Ernst.
\newblock Optimal decision making for complex problems course at the
  {U}niversity of {L}iège.
\newblock
  \url{http://blogs.ulg.ac.be/damien-ernst/info8003-1-optimal-decision-making-for-complex-problems}.
\newblock (Accessed on 05/13/2021).

\end{thebibliography}

\end{document}